\documentclass[10pt,twocolumn,letterpaper]{article}

\usepackage[pagenumbers]{cvpr} 

\usepackage{amsmath}
\usepackage{algorithm}
\usepackage{algorithmic}

\usepackage{pifont}
\newcommand{\cmark}{\ding{51}}%
\newcommand{\xmark}{\ding{55}}%

\definecolor{cvprblue}{rgb}{0.21,0.49,0.74}
\usepackage[pagebackref,breaklinks,colorlinks,allcolors=cvprblue]{hyperref}

\title{Timestep-Aware Block Masking for Efficient Diffusion Model Inference}

\author{
	Haodong He\textsuperscript{1}, 
	Yuan Gao\textsuperscript{2}, 
	Weizhong Zhang\textsuperscript{3}, 
	Gui-Song Xia\textsuperscript{2$\dagger$}\\
	\textsuperscript{1}School of Computer Science, Wuhan University\\
    \textsuperscript{2}School of Artificial Intelligence, Wuhan University\\
    \textsuperscript{3}School of Data Science, Fudan University\\
	{\tt\small \{haodonghe, guisong.xia\}@whu.edu.cn, ethan.y.gao@gmail.com, weizhongzhang@fudan.edu.cn}
}

\begin{document}
\maketitle

\renewcommand{\thefootnote}{}\footnote{$^\dagger$Corresponding author.}
\renewcommand{\thefootnote}{\arabic{footnote}}

\begin{figure*}[h]
    \centering
    \includegraphics[width=\linewidth]{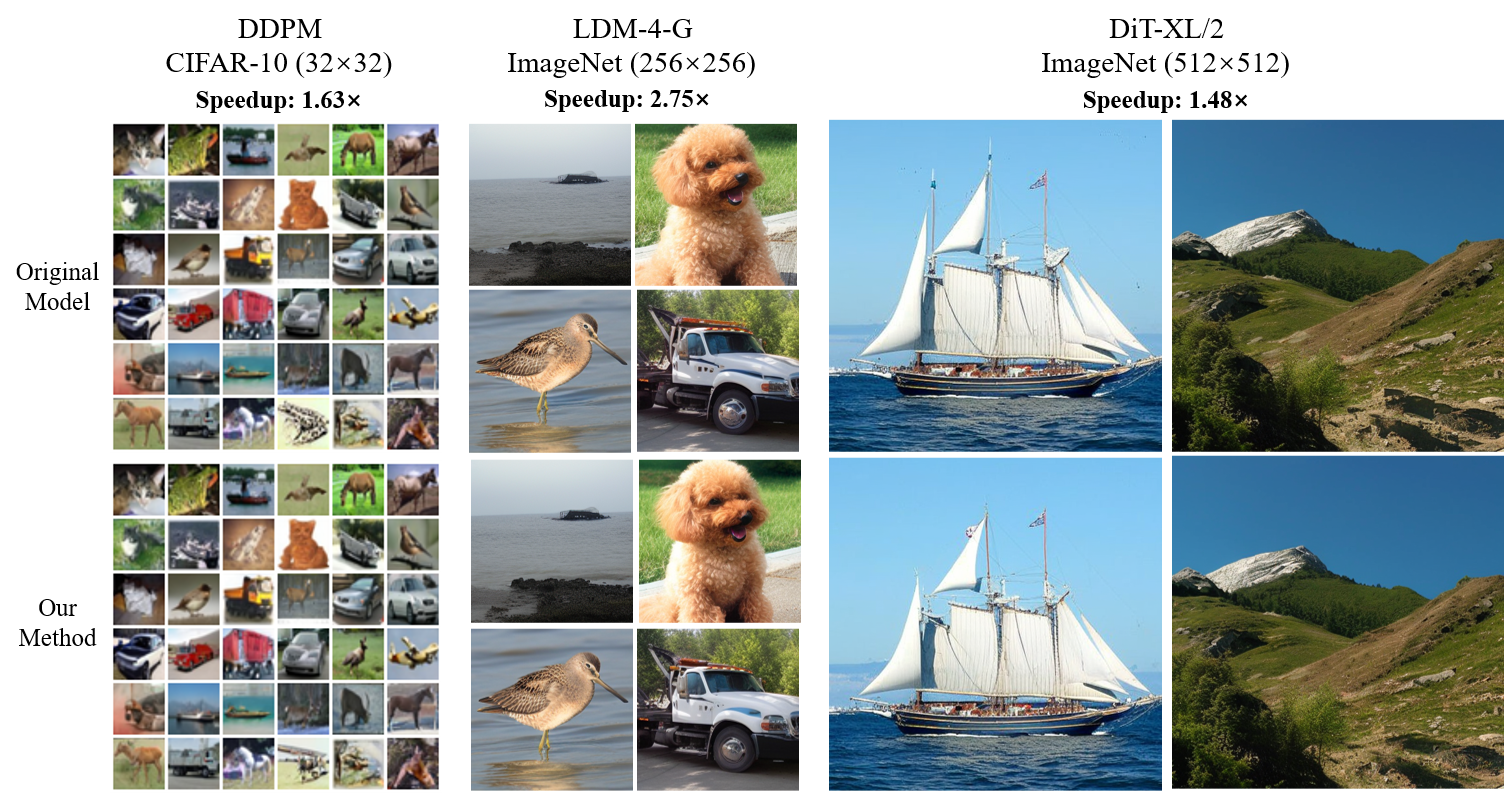}
    \caption{Qualitative demonstration of our method on representative diffusion architectures. In these instances, our method achieves substantial speedups—specifically 1.63$\times$, 2.75$\times$, and 1.48$\times$—while preserving high-fidelity generation quality with high visual consistency.}
    \label{fig:example}
\end{figure*}

\begin{abstract}
Diffusion Probabilistic Models (DPMs) have achieved great success in image generation but suffer from high inference latency due to their iterative denoising nature. Motivated by the evolving feature dynamics across the denoising trajectory, we propose a novel framework to optimize the computational graph of pre-trained DPMs on a per-timestep basis. By learning timestep-specific masks, our method dynamically determines which blocks to execute or bypass through feature reuse at each inference stage. Unlike global optimization methods that incur prohibitive memory costs via full-chain backpropagation, our method optimizes masks for each timestep independently, ensuring a memory-efficient training process. To guide this process, we introduce a timestep-aware loss scaling mechanism that prioritizes feature fidelity during sensitive denoising phases, complemented by a knowledge-guided mask rectification strategy to prune redundant spatial-temporal dependencies. Our approach is architecture-agnostic and demonstrates significant efficiency gains across a broad spectrum of models, including DDPM, LDM, DiT, and PixArt. Experimental results show that by treating the denoising process as a sequence of optimized computational paths, our method achieves a superior balance between sampling speed and generative quality. Our code will be released.
\end{abstract}    
\section{Introduction}

Diffusion Probabilistic Models (DPMs) \cite{sohl2015deep,song2019generative,ho2020ddpm,dhariwal2021adm,rombach2022high,peebles2023scalable} have achieved significant success in various fields such as the generation of images \cite{song2020sliced,vahdat2021score,kawar2022denoising}, speech \cite{huang2022prodiff,huang2022fastdiff}, text \cite{li2022diffusion,gong2022diffuseq}, video \cite{ho2022video,luo2023videofusion}, and 3D objects \cite{chen2023single,cao2023large}, attracting increasing attention. Despite their extraordinary performance, DPMs require the repeated use of denoising models during the inference process to transform Gaussian noise into samples, which usually comes with high computational costs. This trade-off between performance and efficiency poses a critical challenge in resource-constrained environments and has become a bottleneck for the broader adoption of diffusion models \cite{li2024snapfusion}.

Recently, numerous approaches have been proposed to enhance the inference efficiency of diffusion models. They can be broadly categorized into two types: The first explores more efficient sampling strategies. For instance, some studies expedite model inference by refining the solvers \cite{song2020ddim,lu2022dpmsolver,zhang2022fast,zhou2024fast,xue2024accelerating}, whereas others reduce inference timesteps through distillation \cite{salimans2022progressive,meng2023distillation,liu2023instaflow,li2024snapfusion}. The second type accelerates sampling at each timestep: for example, Quantization \cite{li2023q,shang2023post} and pruning \cite{fang2023structural} have also been explored to accelerate model inference. Recent findings \cite{ma2023deepcache,wimbauer2024cache,li2023faster,ma2024learningtocache,selvaraju2024fora} indicate minimal variation in the features output by diffusion models across adjacent timesteps, leading to methods that reuse block outputs for accelerated inference.

Drawing inspiration from these cache-based methods, our goal is to develop a mask for diffusion models that can significantly speed up the model's inference process while preserving its accuracy. Due to the nature of diffusion models, they reuse the denoising network across timesteps, but the impact of each block on image quality varies, allowing selective computation bypass. Specifically, we initialize a mask $\mathbf{m}$ for the denoising model, and the value \( m_{t, b} \) determines the operation that block \( b \) in the model will perform at timestep \( t \): 0 skips the computation step and reuses the cached features; 1 executes the computation step and updates the corresponding cache.

In contrast to rule-based methods like DeepCache \cite{ma2023deepcache} or feature-difference-based methods like the one proposed by Wimbauer \cite{wimbauer2024cache}, the training process is entirely end-to-end, ensuring that the mask is optimal. During training, we will freeze the model's parameters and perform the denoising process, making the outputs of the masked model at each timestep as close as possible to the original model, and constraining the model's efficiency through the $\ell_{1}$ loss of the mask. A critical insight is that the nature of the noise removed by the model varies across timesteps during inference. To address this, we introduce a timestep-aware loss weight to guide the optimization process. After obtaining the mask, we will further perform rectification on it to enhance its ability to accelerate the model without sacrificing accuracy. Unlike L2C \cite{ma2024learningtocache}, the mask's training is tailored around the model's sampling procedure, which means it only requires initializing with Gaussian noise as the input. In contrast to DiP-GO's \cite{zhu2024dip} global end-to-end optimization, our method optimizes masks per timestep. This eliminates the need to retain intermediate features from all timesteps for backpropagation, resulting in a more memory-efficient training process.

Experimental results demonstrate that our method achieves superior acceleration performance. We have experimented with four different structures of diffusion models (DDPM \cite{ho2020ddpm}, LDM \cite{rombach2022high}, DiT \cite{peebles2023scalable}, and PixArt \cite{chen2024pixart}) to demonstrate the effectiveness, efficiency, and universality of our method. For DiT-XL/2 with 50 DDIM steps on ImageNet 512 $\times$ 512, we train the mask on a single GeForce RTX 4090 GPU for less than 3 hours, achieving a 1.48$\times$ acceleration in the sampling process.

In summary, we have proposed a new method for accelerating the inference of diffusion models, which is effective for various architectures. Moreover, its performance is significantly better than that of other cache-based methods. The contributions of our paper include:
\begin{itemize}
    \item
    We introduce a novel approach for training an end-to-end mask for a given diffusion model, which enables the model to skip computations of certain blocks, thereby improving sampling efficiency without requiring retraining of the pre-trained model.
    \item 
    Our method is grounded in the denoising process of the model and requires only initialized noise as input. The training is performed per timestep, making it highly efficient and lightweight. Furthermore, we incorporate timestep-aware loss weighting to guide the optimization and propose a mask post-processing technique to further enhance acceleration.
    \item 
    Our method is universal and effective across various architectures, such as DDPM, LDM, DiT, and PixArt on CIFAR-10, LSUN-Bedroom, LSUN-Churches, ImageNet, and MS-COCO.
\end{itemize}
\section{Related Work}
\paragraph{Diffusion Models.} In the field of deep learning, diffusion models have emerged as a novel class of generative models that have undergone rapid development after GANs \cite{goodfellow2014gan,arjovsky2017wassersteingan} and VAEs \cite{kingma2013vae,higgins2016betavae}. They have demonstrated exceptional performance in multiple domains, including image generation \cite{song2020sliced,vahdat2021score,kawar2022denoising}, video synthesis \cite{ho2022video,luo2023videofusion}, 3D objects modeling \cite{chen2023single,cao2023large}, and so on. DDPM \cite{ho2020ddpm} represents pioneering work in diffusion models, where noise is gradually added to the pixel space and then a neural network is trained to reverse this process, generating high-quality images. LDM \cite{rombach2022high} builds upon DDPM by conducting the diffusion process in the latent space rather than the pixel space, significantly reducing computational complexity while maintaining the quality of generated images. The recently introduced DiT \cite{peebles2023scalable} combines diffusion models with the Transformer \cite{vaswani2017attention} architecture, leveraging the powerful modeling capabilities of Transformers to handle the latent representation of images. While diffusion models exhibit superior performance, the repetitive application of the underlying denoising neural networks renders them computationally expensive. To further the democratization of diffusion models, this work focuses on accelerating the sampling speed of these models, ensuring precision is maintained.

\paragraph{Model Acceleration.} The acceleration of diffusion models can generally be divided into two categories. The first category involves reducing the number of time steps required for model sampling. Representative approaches in this domain include various enhanced solvers. For instance, DDIM \cite{song2020ddim} reduces time steps by exploring a non-Markovian process, which is related to neural ODEs. Subsequently, numerous studies have been proposed focusing on fast solvers of SDEs or ODEs to enable efficient sampling \cite{lu2022dpmsolver,zhang2022fast,zhou2024fast,xue2024accelerating}. Furthermore, some research has aimed to decrease sampling steps through distillation \cite{salimans2022progressive,meng2023distillation,liu2023instaflow,li2024snapfusion}. Consistency models \cite{song2023consistency,luo2023latent} can even generate high-quality images in few steps. The second category focuses on reducing the computational time per individual denoising step. Many studies have attempted to speed up model inference through pruning \cite{fang2023structural} or quantization \cite{li2023q,shang2023post} methods. 

In addition, some research \cite{ma2023deepcache,wimbauer2024cache,li2023faster,ma2024learningtocache,selvaraju2024fora} has achieved model acceleration through block caching. For example, DeepCache \cite{ma2023deepcache} has introduced a rule-based model acceleration method without training, while Wimbauer \cite{wimbauer2024cache} determines whether to skip a block by assessing the differences. Furthermore, L2C \cite{ma2024learningtocache} models the diffusion model's training process and optimizes a router to determine which modules to skip. In contrast, our method, similar to DiP-GO \cite{zhu2024dip}, is modeled based on the sampling process. The key difference lies in our per-timestep optimization approach, which requires less GPU memory during training while achieving superior performance. 

\section{Method}

\subsection{Preliminary}
\paragraph{Diffusion models.} Diffusion models are a cutting-edge class of generative models that simulate the data generation process. They involve a forward diffusion process, where data gradually evolves into noise, and a reverse diffusion process, where noise is incrementally transformed back into data. The forward diffusion can be summarized as:
\begin{equation}
x_{t}=\sqrt{\alpha_{t}} x_{t-1}+\sqrt{1-\alpha_{t}} z_{t},
\label{diffusion_equation_1}
\end{equation}where $x_t$ is the data at step $t$, $z_t$ is Gaussian noise, and $\alpha_{t}$ is a predefined sequence of noise levels. The reverse diffusion, which is the generative phase, is given by:
\begin{equation}
x_{t-1}=\frac{1}{\sqrt{\alpha_{t}}}\left(x_{t}-\frac{1-\alpha_{t}}{\sqrt{1-\bar{\alpha}_{t}}} \epsilon_{\theta}\left(x_{t}, t\right)\right),
\label{diffusion_equation_2}
\end{equation}where $\epsilon_{\theta}\left(x_{t}, t\right)$ is the model's prediction of the noise at step $t$, and $\bar{\alpha}_{t}$ is the cumulative product of $\alpha$ up to $t$.

\paragraph{Block caching.} Block caching is a strategy to accelerate the inference speed of diffusion models. Specifically, since previous research has demonstrated that the U-Net \cite{ronneberger2015unet} network of the diffusion model exhibits minimal changes in underlying features across adjacent timesteps during inference, the outputs of certain blocks at the current timestep can be cached and reused by the corresponding blocks at the next timestep. This allows for skipping the computation of some blocks, thereby accelerating the model's inference.

\subsection{Problem Formulation}
We aim to learn a two-dimensional  binary mask $\mathbf{m}$, where its element \( m_{t, b} \) determines whether the block \( b \) of the model at timestep \( t \) should be skipped (i.e., \( m_{t, b} = 0 \)) or not (i.e., \( m_{t, b} = 1 \)). The binary mask $\mathbf{m}$ forms a $T \times B$ matrix, where $T$ is the total sampling timestep and $B$ is the total number of blocks of the network architecture. The flexible mask formulation enables skipping the network blocks in various granularities, and in our implementation, we consider the \emph{Multi-Head Attention (MHA)} and \emph{MLP} for the Diffusion Transformer (DiT) architecture, as well as \emph{ResBlock} and \emph{AttnBlock} for the U-Net CNN architecture.

During the training process, we freeze the model parameters and only train the mask $\mathbf{m}$, greatly reducing the demand for GPU memory. For each block $b$, we cache its feature and determine whether the cached features are reusable at timestep $t$ by \( m_{t, b} \).

\begin{figure*}[t]
    \centering
    \includegraphics[width=\linewidth]{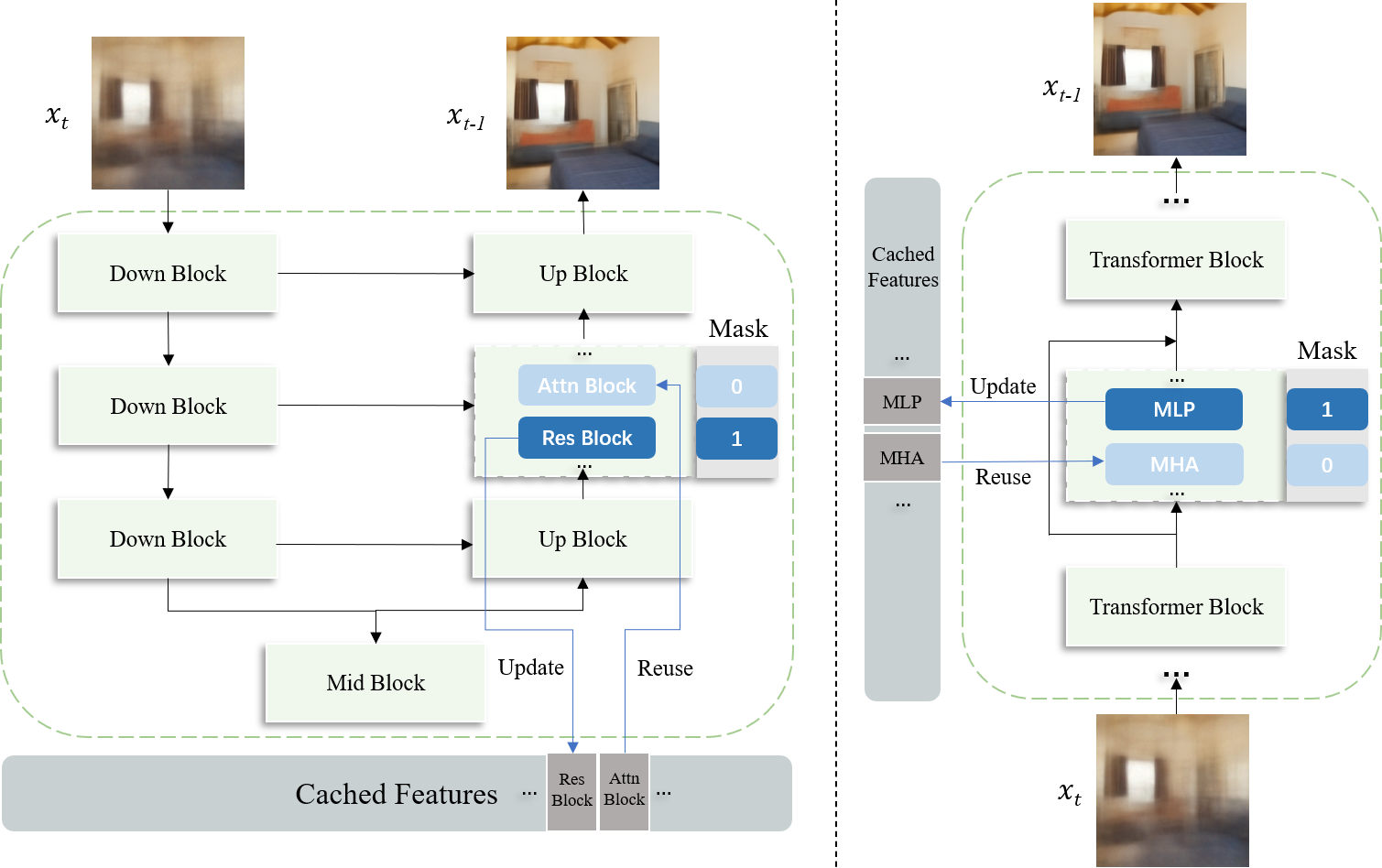}
    \caption{Illustration of our method on a UNet-based model (left) and a Transformer-based model (right). During the denoising phase, upon receiving the input $x_t$, the model checks the mask $m_t$ values corresponding to each block. If the mask value is 1, it performs computations and updates the cached features for that block; if the mask value is 0, it skips the computation and uses the existing cached features.}
    \label{fig:block_mask}
    \vspace{-2mm}
\end{figure*}

Formally, for each block $b$ at timestep \(t\), we learn a binary mask $m_{t, b}$ to determine whether its feature $x_{t, b}$ should be computed through the current block, \ie, $f_{t, b}(x_{t, b-1})$ where $f_{t, b}$ is the current block such as \emph{MHA, MLP, ResBlock}, or \emph{AttnBlock} and $x_{t, b-1}$ is its input; or simply reuse the cached feature $x^{\text{cache}}_{b}$ from the previous timestep for acceleration. Our cached feature for each block is continuously updated as the timestep increases. The above process is illustrated in Fig. \ref{fig:block_mask}, and this gives us the following formulation:
\begin{equation}
\begin{aligned}
& x_{t, b} = m_{t, b} f_{t, b}(x_{t, b-1}) + (1 - m_{t, b}) x^{\text{cache}}_{b},
\\ &x^{\text{cache}}_{b} = x_{t, b}, \text{ if } m_{t, b} = 1 \qquad \text{s.t.} \quad m_{t, b} \in\{0,1\}.
\end{aligned}
\label{eq:mask1}
\end{equation}

\subsection{Optimization}
We observe that the binary optimization problem, which involves finding a high-quality mask $\mathbf{m}$ within a discrete and exponentially expanding solution space, is generally NP-hard. Thus, we need to transform the discrete elements \( m_{t, b} \) in the mask into continuous variables \( s_{t, b} \). However, when considering the use of Gumbel-Softmax sampling, as the temperature parameter increases, its distribution tends to approximate uniform sampling, which introduces bias. And we cannot directly control its variance. Hence, we choose to perform continuous random sampling of the mask between 0 and 1, setting the probability of each element \( m_{t, b} \) to be 1 as \( s_{t, b} \) and the probability to be 0 as \( 1 - s_{t, b} \). Then, We use regularization terms to encourage these values to converge to either 0 or 1. 

Equation \ref{eq:mask1} has thus transformed into:
\begin{equation}
\begin{aligned}
& x_{t, b} = s_{t, b} f_{t, b}(x_{t, b-1}) + (1 - s_{t, b}) x^{\text{cache}}_{b},
\\ &x^{\text{cache}}_{b} = x_{t, b}, \text{ if } s_{t, b} > 0.5 \text{ or } t = T-1 \\ & \qquad\qquad \text{s.t.}  \quad s_{t, b} \in[0,1].
\end{aligned}
\label{eq:mask2}
\end{equation}

\begin{figure}[h]
    \centering
    \includegraphics[width=\linewidth]{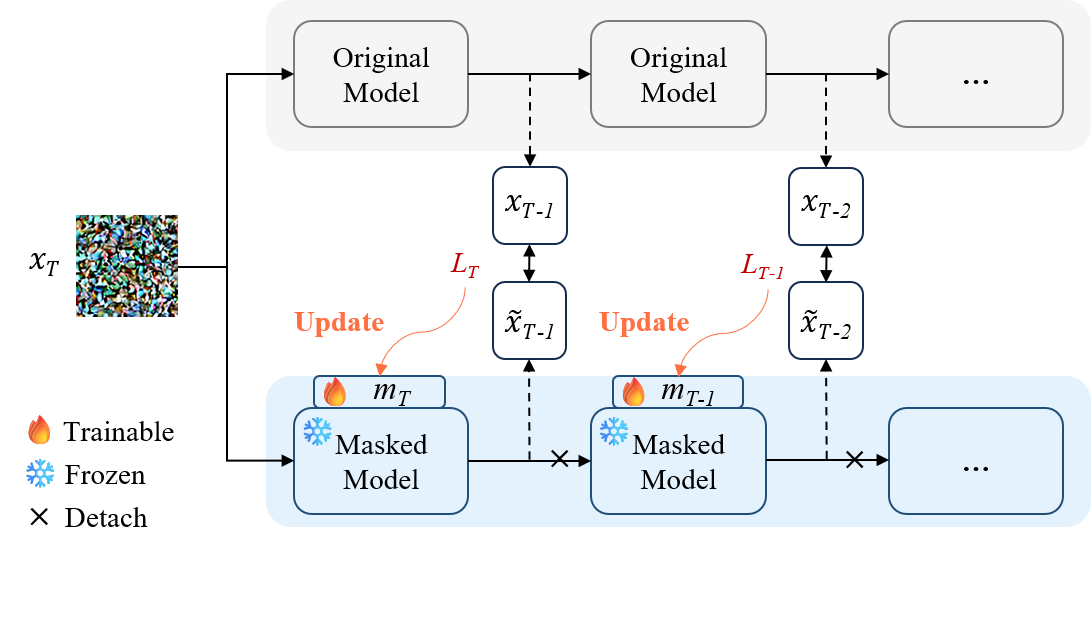}
    \caption{Illustration of the end-to-end mask optimization framework. The original DPM is frozen, serving as a teacher to provide reference features $x_t$. For each timestep, the learnable mask $m_t$ is updated by the total loss $L_t$. The \textit{Detach} operation ensures that gradients are restricted to the current timestep, enabling memory-efficient per-step optimization.}
    \label{fig:mask_train}
\end{figure}

As illustrated in Fig. \ref{fig:mask_train}, we restrict our optimized feature $x_{t, end}$ output by the end block under the learned mask $m_{t,b}$ to introduce minimal distortion of the original feature $x^{\text{ori}}_{t, end}$ from the vanilla DPM, which yields the following $\ell_2$ feature loss for each timestep $t$:
\begin{equation}
L^{\text{feature}}_t = || x_{t, end} - x^{\text{ori}}_{t, end}||_2.
\label{eq:mask3}
\end{equation}

Simultaneously, we would like to learn a sparse $\mathbf{m}$ to identify and skip all the less important blocks, indicating the $\ell_{1}$ regularization on $m_{t,b}$ as:
\begin{equation}
L^{\text{sparse}}_t = \sum_b || s_{t, b} ||_1.
\label{eq:mask4}
\end{equation}

Additionally, to encourage the elements of \( m_t \) to converge towards binary values of 0 and 1, we introduce a bi-modal regularizer \cite{srinivas2017training}:

\begin{equation}
L^{\text{bi-modal}}_t = \sum_b s_{t,b}(1-s_{t,b}).
\label{eq:mask5}
\end{equation}

From Eq. \ref{eq:mask3}, \ref{eq:mask4}, and \ref{eq:mask5}, it can be derived that our optimization objective is as follows:

\begin{equation}
L_t = L^{\text{feature}}_t + \lambda_1 L^{\text{sparse}}_t + \lambda_2 L^{\text{bi-modal}}_{t}.
\label{eq:mask6}
\end{equation}

In which \(\lambda_1\) and \(\lambda_2\) are the regularization coefficients for the two regularization terms, respectively. 

\subsection{Timestep-Aware Loss Scaling}

Due to the varying noise levels processed by the model at different timesteps during inference, we introduce a timestep-aware loss weight to guide the optimization process. Specifically, smaller regularization coefficients are applied at critical timesteps with significant feature changes to prioritize generation quality, while larger regularization coefficients are used at timesteps with smooth feature transitions to aggressively enhance acceleration.

Specifically, we compute the relative feature variation at each timestep using the pre-trained model:
\begin{equation}
\delta[t] = \frac{|| x^{\text{ori}}_{t, end} - x^{\text{ori}}_{t-1, end}||_2}{||x^{\text{ori}}_{t, end}||_2}.
\end{equation}

A piecewise function is then employed to assign the loss weight:
\begin{equation}
w(t) =
\begin{cases}
2.0, & \delta[t]/\max(\delta) < 0.1 \\
1.5, & 0.1 \leq \delta[t]/\max(\delta) < 0.5 \\
1.0, & \text{otherwise}
\end{cases}
\end{equation}

The final loss function for timestep is defined as:
\begin{equation}
L_t = L^{\text{feature}}_t + \lambda_1 w(t) L^{\text{sparse}}_t + \lambda_2 w(t) L^{\text{bi-modal}}_{t}.
\end{equation}

\subsection{Knowledge-Guided Mask Rectification}

Apart from optimizing $m_{t, b}$ to 0 by using the sparse regularization Eq. \ref{eq:mask4}, there exists an additional postprocessing rule to \emph{safely} rectify $m_{t, b}$ to 0 after optimization, resulting in a further acceleration in the inference. Such a rectification rule is derived leveraging the mask dependencies 1) \emph{among different blocks within the same timestep} and 2) \emph{between the same blocks from the adjacency timesteps}, \emph{without additional training}.

Specifically, by iterating Eq. \ref{eq:mask2} from the first timestep and block to the last one, a certain feature $x_{t, b}$ can be used in two ways: 1) acting as the input of \emph{its next block within the same timestep}, and 2) potentially being reused by \emph{the same block of the next timestep}. If both blocks do not need $x_{t, b}$ as input, we can \emph{safely} set $m_{t, b}$ to 0.

Considering block $b$ at timestep $t$, if its next block $b+1$ at timestep $t$ reuses the cached features, \ie, $m_{t, b+1} = 0$, then we no longer need $x_{t, b}$ as input to calculate $x_{t, b+1}$. 
Simultaneously, if the same block $b$ at its next timestep $t-1$ does not reuse $x_{t, b}$ as cache, \ie, $m_{t-1, b} = 1$, we can safely skip block $b$ at timestep $t$ without additional training:
\begin{equation}
m_{t, b} = 0, \quad \text{~ if ~} m_{t, b+1} = 0 \textbf{~ and ~} m_{t-1, b} = 1.
\label{eq:rectification}
\end{equation}

\begin{figure}[h]
    \centering
    \includegraphics[width=\linewidth]{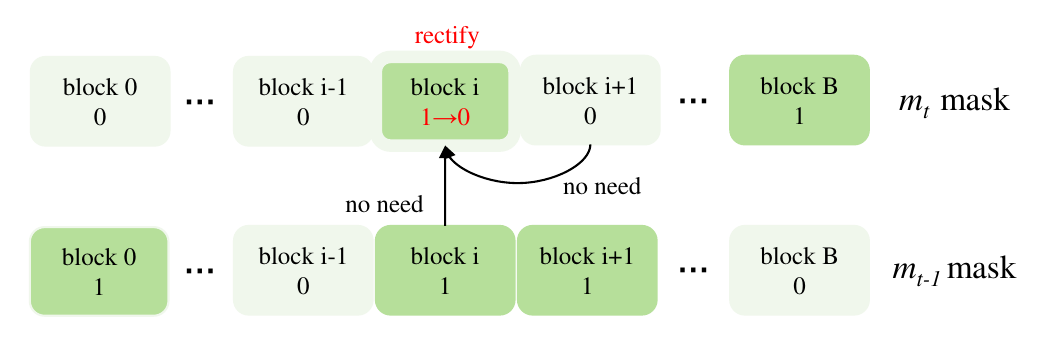}
    \caption{Illustration of mask rectification. Within the same timestep, if the subsequent block does not require its computation result as input, and the same block in the next timestep does not need to reuse its features, then the computation of this block can be safely skipped.}
    \label{fig:rectification}
    \vspace{-1mm}
\end{figure}

Finally, we use Eq. \ref{eq:rectification} to rectify our masks from timestep 0 (\ie, the last timestep) to timestep $T$ (\ie, the first timestep), and within each timestep, from the last block to the first one. The above process is illustrated in Fig. \ref{fig:rectification}.

\section{Experiments}

\begin{figure*}[h]
    \centering
    \includegraphics[width=0.9\linewidth]{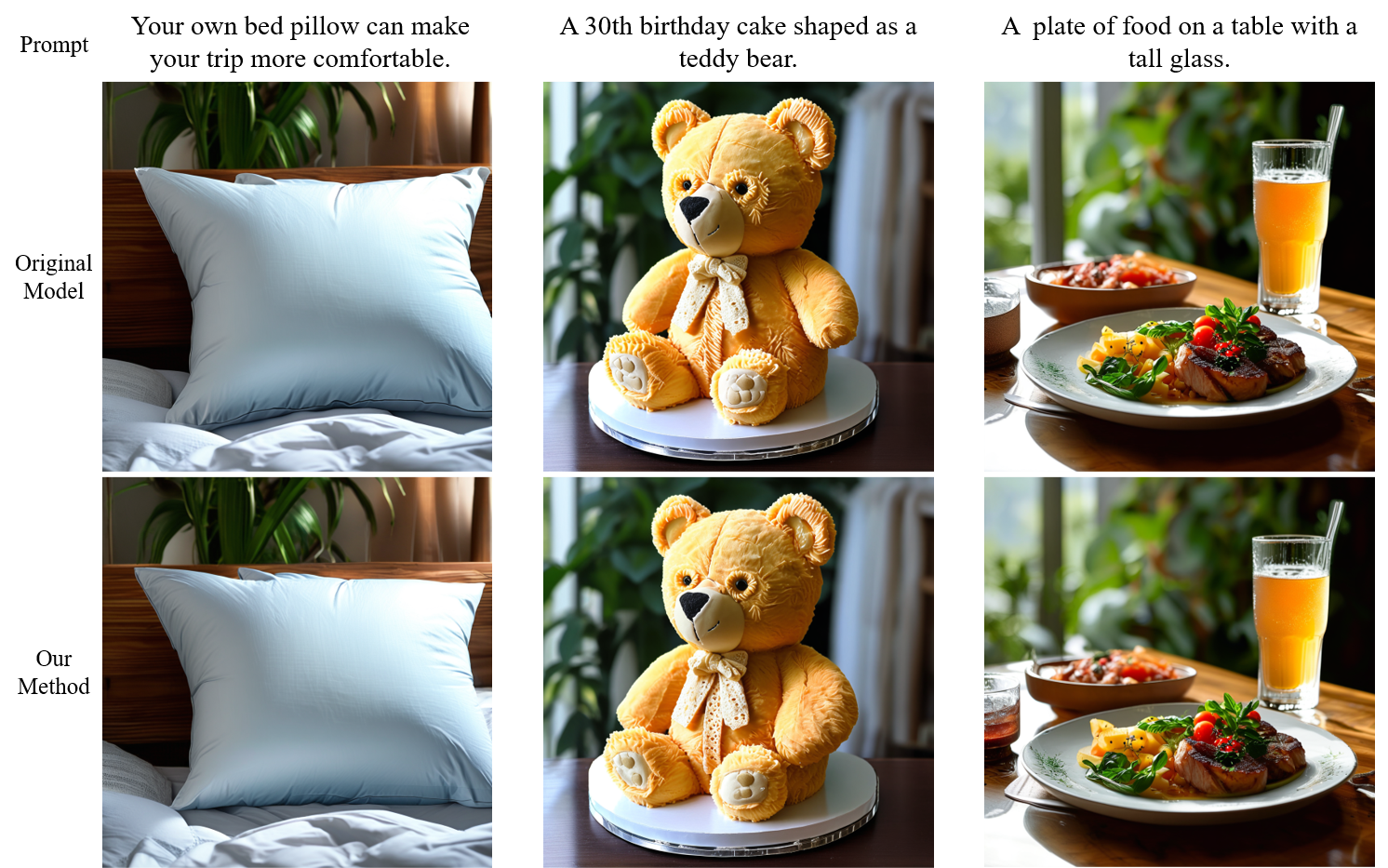}
    \caption{Acceleration for PixArt-Sigma-XL on MS-COCO 1024 $\times$ 1024. Our method not only achieves a 1.43 $\times$ speedup, but also maintains excellent image generation quality and text-to-image semantic consistency.}
    \label{fig:visualization}
\end{figure*}

\subsection{Experimental Setup}

\paragraph{Models.}

To demonstrate the effectiveness and universality of our approach, we conduct experiments on DDPM \cite{ho2020ddpm}, LDM \cite{rombach2022high}, DiT \cite{peebles2023scalable}, and PixArt \cite{chen2024pixart}. We select pre-trained models DDPM, LDM-4-G, DiT-XL/2, and PixArt-Sigma-XL for subsequent mask training.

\paragraph{Datasets.}

Our evaluations are conducted on multiple standard benchmarks across diverse resolutions. For the CNN-based DDPM, we utilize CIFAR-10~\cite{krizhevsky2009cifar10} ($32\times32$) and LSUN~\cite{yu2015lsun} ($256\times256$, Bedroom and Churches). Regarding ImageNet~\cite{deng2009imagenet}, we adopt the $256\times256$ version for LDM-4-G, while evaluating DiT-XL/2 at both $256\times256$ and $512\times512$ scales. For PixArt-Sigma-XL, MS-COCO~\cite{lin2014mscoco} is employed at a resolution of $1024\times1024$.

\paragraph{Baselines.}

For DDPM and LDM-4-G experiments, we choose DeepCache \cite{ma2023deepcache} as the primary baseline for comparison, which has achieved excellent results as a training-free diffusion model acceleration method. For DiT-XL/2 experiments, we select Learning-to-Cache (L2C) \cite{ma2024learningtocache} as the primary baseline, which trains a router based on the model's process of adding noise to skip the computation, requiring additional training data as input. For PixArt-Sigma-XL experiments, we choose DitFastAttn \cite{yuan2024ditfastattn}, a post-training compression method, as the primary baseline. 

\paragraph{Evaluation Metrics.}

For DDPM experiments, we use FID~\cite{heusel2017fid}  to evaluate the quality of the generated images. For LDM-4-G and DiT-XL/2 experiments, in addition to FID, we also considered sFID~\cite{nash2021sFID}, IS~\cite{salimans2016improved}, Precision, and Recall as metrics. For PixArt-Sigma-XL experiments, IS, FID, and CLIP Score~\cite{hessel2021clipscore} are used to evaluate semantic alignment and visual fidelity. We use inference speed and total MACs as metrics to quantify the computational efficiency and acceleration of the methods.

\begin{table}[h]
\centering
    \small
    \resizebox{\linewidth}{!}{
    \begin{tabular}{l | c c c c | c}
      \toprule
      \multicolumn{6}{c}{\it CIFAR-10 32 $\times$ 32 } \\
      \bf Method & \bf Extra Data & \bf Training Time $\downarrow$ & \bf MACs $\downarrow$ & \bf Speed $\uparrow$ & \bf FID $\downarrow$ \\
      \midrule
      DDPM \cite{ho2020ddpm}           & -- & -- & 0.61T & 1$\times$ & 4.19 \\
      DDPM*                             & -- & -- & 0.61T & 1$\times$ & 4.25 \\
      \cmidrule(lr){1-6}
      Diff-Pruning \cite{fang2023structural} & \cmark & -- & 0.34T & 1.37$\times$ & 5.29 \\
      CT \cite{song2023consistency}*   & \cmark & -- & --    & 1.62$\times$ & 4.68 \\
      DeepCache \cite{ma2023deepcache} & \xmark & \xmark & 0.35T & 1.61$\times$ & 4.70 \\
      \bf Ours                         & \xmark & 0.2h & 0.34T & \bf 1.63$\times$ & \bf 4.66 \\
      
      \midrule
      \multicolumn{6}{c}{\it LSUN-Bedroom  256 $\times$ 256}\\
      \bf Method & \bf Extra Data & \bf Training Time $\downarrow$ & \bf MACs $\downarrow$ & \bf Speed $\uparrow$ & \bf FID $\downarrow$ \\
      \midrule
      DDPM \cite{ho2020ddpm}           & -- & -- & 24.9T & 1$\times$ & 6.62 \\
      DDPM*                             & -- & -- & 24.9T & 1$\times$ & 6.75 \\
      \cmidrule(lr){1-6}
      Diff-Pruning \cite{fang2023structural} & \cmark & -- & 13.9T & \bf 1.48$\times$ & 18.60 \\
      DeepCache \cite{ma2023deepcache} & \xmark & \xmark & 19.1T & 1.29$\times$ & 6.69 \\
      \bf Ours                         & \xmark & 0.6h & 18.3T & 1.34$\times$ & \bf 6.67 \\
      
      \midrule
      \multicolumn{6}{c}{\it LSUN-Churches  256 $\times$ 256 } \\
      \bf Method & \bf Extra Data & \bf Training Time $\downarrow$ & \bf MACs $\downarrow$ & \bf Speed $\uparrow$ & \bf FID $\downarrow$ \\
      \midrule
      DDPM \cite{ho2020ddpm}           & -- & -- & 24.9T & 1$\times$ & 10.58 \\
      DDPM*                             & -- & -- & 24.9T & 1$\times$ & 10.92 \\
      \cmidrule(lr){1-6}
      Diff-Pruning \cite{fang2023structural} & \cmark & -- & 13.9T & \bf 1.48$\times$ & 13.90 \\
      DeepCache \cite{ma2023deepcache} & \xmark & \xmark & 19.1T & 1.29$\times$ & 11.31 \\
      \bf Ours                         & \xmark & 0.6h & 18.8T & 1.31$\times$ & \bf 10.39 \\
      \bottomrule
    \end{tabular}
    }
    \caption{Unconditional generation quality using DDPM on CIFAR-10, LSUN-Bedroom, and LSUN-Churches. All the methods here adopt 100 DDIM steps, except for CT. CT means consistency model (with 55 denoising steps). * means the reproduced results. \textbf{Bold} indicates the best performance.}
    \vspace{-2mm}
    \label{tbl:ddpm}
\end{table}

\begin{table*}[h]
\centering
    \small
    \resizebox{\linewidth}{!}{
    \begin{tabular}{l | c | c c c c | c c c c c }
      \toprule
      \multicolumn{11}{c}{\it ImageNet 256 $\times$ 256}  \\
      \bf Method & \bf NFE & \bf Extra Data & \bf Training Time $\downarrow$ & \bf MACs $\downarrow$ & \bf Speed $\uparrow$ & \bf IS $\uparrow$ & \bf FID $\downarrow$ & \bf sFID $\downarrow$ & \bf Precision $\uparrow$& \bf Recall $\uparrow$  \\
      \midrule
      LDM-4~\cite{rombach2022high}* & 250 & -- & -- & 25.0T & 1$\times$ & 206.45 & 3.42 & 5.14 & 82.83 & 53.13 \\
\cmidrule(lr){1-11}
Diff-Pruning~\cite{fang2023structural} & 250 & \cmark & -- & 13.2T & 1.51$\times$ & 201.81 & 9.16 & 10.59 & \bf 87.87 & 30.87 \\
DeepCache~\cite{ma2023deepcache} & 250 & \xmark & \xmark & 9.1T & 2.65$\times$ & 202.79 & \bf 3.44 & 5.11 & 82.65 & \bf 53.81 \\
\bf Ours & 250 & \xmark & 2.9h & 8.3T & \bf 2.75$\times$ & \bf 205.76 & 3.51 & \bf 5.00 & 82.74 & 52.81 \\
\midrule
DiT-XL/2~\cite{peebles2023scalable}* & 100 & -- & -- & 11.9T & 1$\times$ & 242.8 & 2.16 & 4.45 & 80.35 & 60.34 \\
\cmidrule(lr){1-11}
L2C~\cite{ma2024learningtocache}* & 100 & \cmark & 17.0h & 8.1T & 1.38$\times$ & \bf 240.7 & \bf 2.29 & 4.52 & \bf 80.30 & \bf 60.04 \\
FORA~\cite{selvaraju2024fora}* & 100 & \xmark & \xmark & -- & \bf 1.74$\times$ & 232.7 & 4.29 & 8.59 & 76.46 & 58.11 \\
DiP-GO~\cite{zhu2024dip} & 250 & \xmark & -- & 7.4T & 1.46$\times$ & -- & 3.14 & -- & -- & -- \\
\bf Ours & 100 & \xmark & 2.5h & 6.7T & 1.67$\times$ & 240.2 & \bf 2.29 & \bf 4.45 & 80.00 & \bf 60.04 \\
\midrule

      \multicolumn{11}{c}{\it ImageNet 512 $\times$ 512}  \\
      \bf Method & \bf NFE & \bf Extra Data  & \bf Training Time $\downarrow$ & \bf MACs $\downarrow$ & \bf Speed $\uparrow$ & \bf IS $\uparrow$ & \bf FID $\downarrow$ & \bf sFID $\downarrow$ & \bf Precision $\uparrow$& \bf Recall $\uparrow$  \\
      \midrule
      
      DiT-XL/2~\cite{peebles2023scalable}* & 50 & -- & -- & 26.3T & 1$\times$ & 203.5 & 3.33 & 4.53 & 83.42 & 54.50 \\
\cmidrule(lr){1-11}
L2C~\cite{ma2024learningtocache}* & 50 & \cmark & 18.4h & 16.9T & 1.54$\times$ & 200.7 & 3.76 & 5.10 & 83.16 & 54.30 \\
FORA~\cite{selvaraju2024fora}* & 50 & \xmark & \xmark & -- & \bf 1.67$\times$ & 75.7 & 42.43 & 21.04 & 56.06 & 49.60 \\
\bf Ours & 50 & \xmark & 2.8h & 17.9T & 1.48$\times$ & \bf 202.8 & \bf 3.64 & \bf 5.06 & \bf 83.18 & \bf 54.50 \\
      \bottomrule
    \end{tabular}
    }
    \caption{Class-conditional generation quality on ImageNet using LDM-4-G and DiT-XL/2. The baselines here, as well as our methods, employ the DDIM scheduler. * means the reproduced results. \textbf{Bold} indicates the best performance.}
    \label{tbl:ldm_and_dit}
\end{table*}

\begin{table*}[h]
\centering
    \small
    \resizebox{\linewidth}{!}{
    \begin{tabular}{l | c | c c c c | c c c }
      \toprule
      \bf Method & \bf NFE & \bf Extra Data  & \bf Training Time $\downarrow$ & \bf MACs $\downarrow$ & \bf Speed $\uparrow$ & \bf IS $\uparrow$ & \bf FID $\downarrow$ & \bf CLIP Score $\uparrow$ \\
      \midrule
      
      PixArt-Sigma-XL~\cite{chen2024pixart}* & 50 & -- & -- & 107.8T & 1$\times$ & 55.65 & 24.33 & 31.27 \\
\midrule
PixArt-Sigma-XL & 38 & \xmark & \xmark & 81.9T & 1.32$\times$ & 50.88 & 24.89 & 30.02 \\
DitFastAttn~\cite{yuan2024ditfastattn} & 50 & \xmark & 0.3h & -- & 1.24$\times$ & 52.74 & 23.94 & 31.18 \\
\bf Ours & 50 & \xmark & 4.2h & 71.9T & \bf 1.43$\times$ & \bf 55.52 & \bf 22.97 & \bf 31.45 \\
      
      \bottomrule
    \end{tabular}
    }
    \caption{Prompt-conditional generation quality on MS-COCO 1024 $\times$ 1024 using PixArt-Sigma-XL. The baselines here, as well as our methods, employ the DPM-Solver scheduler. * means the reproduced results. \textbf{Bold} indicates the best performance.}
    \vspace{-2mm}
    \label{tbl:pixart}
\end{table*}

\subsection{Main Results}

To demonstrate that our method is effective across different architectures of diffusion models, we conduct experiments on DDPM \cite{ho2020ddpm}, LDM \cite{rombach2022high}, DiT \cite{peebles2023scalable}, and PixArt \cite{chen2024pixart}. The experimental results for DDPM on CIFAR-10 \cite{krizhevsky2009cifar10}, LSUN-Bedroom \cite{yu2015lsun}, and LSUN-Churches \cite{yu2015lsun} are shown in Tab. \ref{tbl:ddpm}. Compared with DeepCache \cite{ma2023deepcache}, we achieve a higher acceleration ratio while maintaining a better FID score on all these datasets. Compared to Diff-Pruning \cite{fang2023structural}, our method does not require additional training data and causes a much lower accuracy loss. Especially on LSUN-Churches, we not only achieved a 1.31$\times$ speedup but also obtained a better FID score compared to the original model.

The experimental results for LDM-4-G \cite{rombach2022high} on ImageNet \cite{deng2009imagenet} are shown in Tab. \ref{tbl:ldm_and_dit}. Our method is capable of accelerating the model to 2.75$\times$ its original speed with minimal performance loss. Compared to DeepCache, we achieve a higher acceleration ratio while outperforming the IS, sFID, and Precision metrics.

The experimental results for DiT-XL/2 \cite{peebles2023scalable} on ImageNet are shown in Tab. \ref{tbl:ldm_and_dit}. Compared with L2C \cite{ma2024learningtocache}, our method only requires initialized Gaussian noise as input and significantly reduces training time. It matches L2C's accuracy but accelerates the model more obviously (1.67$\times$ vs. 1.38$\times$) on ImageNet 256 $\times$ 256. On ImageNet 512 $\times$ 512, our method is slightly slower than L2C but more accurate. Compared with FORA \cite{selvaraju2024fora}, our method is also more advantageous. Although FORA is training-free and achieves the most significant acceleration of the model, its negative impact on the model's accuracy is notable.

Tab. \ref{tbl:pixart} presents the experimental results of PixArt-Sigma-XL \cite{peebles2023scalable} on MS-COCO. Compared with DitFastAttn \cite{ma2024learningtocache} and methods using fewer DPM-Solver steps, our approach achieves superior image generation quality while delivering a higher acceleration ratio. Qualitative comparison results on the model are presented in Fig. \ref{fig:visualization}, and more results can be seen in the supplementary material.

\begin{figure}[t]
    \centering
    \includegraphics[width=\linewidth]{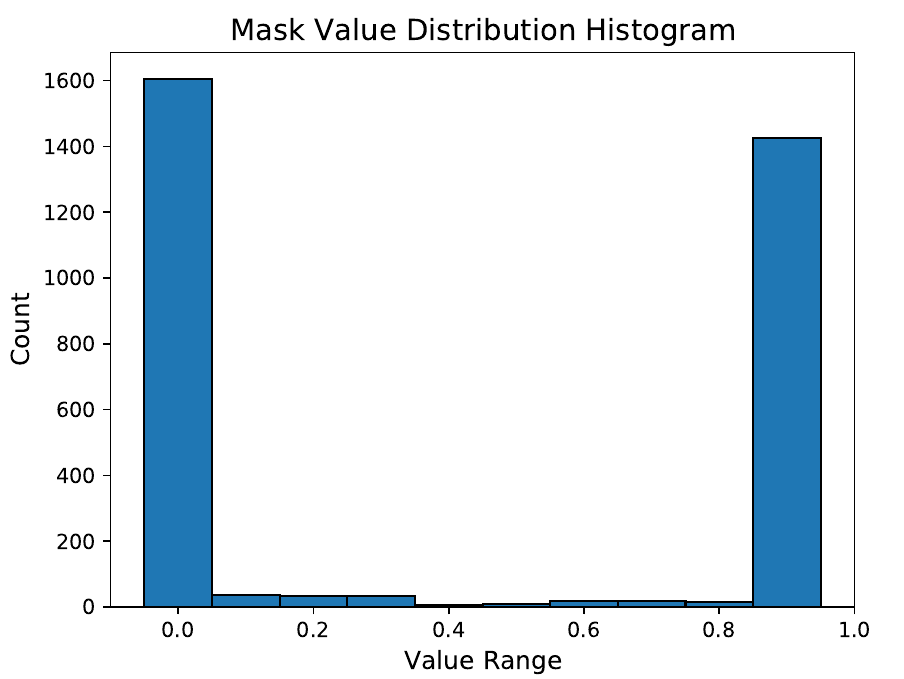}
    \caption{The visualization results of the mask's values show that most elements have converged to near 0 or 1.}
    \label{fig:histogram}
\end{figure}

In Fig. \ref{fig:block_heatmap}, we illustrate the distribution of element values in the mask trained on the CIFAR-10 dataset for DDPM (with 100 DDIM steps) using our method. As shown in it, the mask values exhibit an uneven distribution. Additionally, we conduct a statistical analysis of the value range of elements in the masks, as illustrated in Fig. \ref{fig:histogram}. Most elements converge around 0 and 1.

\begin{figure}[t]
    \centering
    \includegraphics[width=\linewidth]{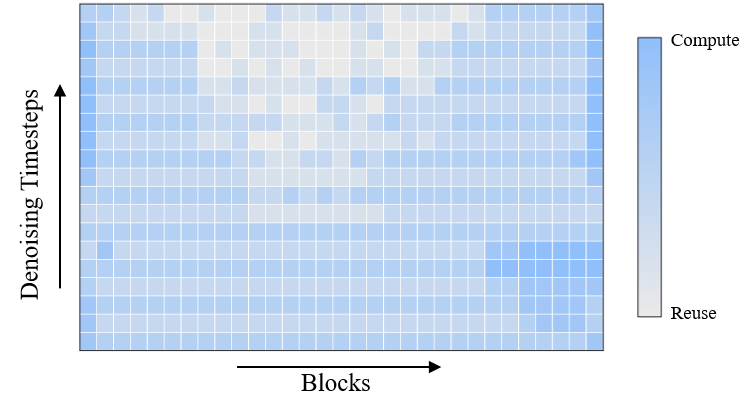}
    \caption{The visualization results of the mask trained by our method. The color of each rectangle represents the number of times the corresponding block performs computations over 5 timesteps; the darker the color, the more computations.}
    \label{fig:block_heatmap}
\end{figure}

\subsection{Ablation Study}

\paragraph{Ablation of mask sampling methods.} We compared the impact of two sampling methodologies (i.e., random sampling and Gumbel-Softmax sampling) on the experimental outcomes, as detailed in Tab. \ref{tbl:ablation_2}. We can observe that the mask derived from training with Gumbel-Softmax sampling performs less effectively in both accelerating the model and maintaining generation accuracy compared to the mask obtained through random sampling. We believe that for Gumbel-Softmax sampling, the high temperature parameter leads to a sampling distribution that more closely approximates a uniform distribution, thereby introducing bias. Moreover, the characteristics of Gumbel-Softmax sampling make it difficult to directly control its variance, limiting the fine-tuning of the optimization process.

\begin{table}[h]
\centering
    \small
    \resizebox{0.6\linewidth}{!}{
    \begin{tabular}{c | c c}
      \toprule
      \multicolumn{3}{c}{\bf CIFAR-10 32 $\times$ 32 } \\
      \midrule
      \bf Sampling Method & \bf Speed $\uparrow$ & \bf FID $\downarrow$ \\
      \midrule
      Random  & 1.63$\times$ & 4.66\\
      Gumbel  & 1.29$\times$ & 4.67\\
      \bottomrule
    \end{tabular}
    }
    \caption{Performance comparison with random sampling and Gumbel-Softmax sampling.}
    \label{tbl:ablation_2}
\end{table}

\begin{table}[h]
\centering
    \small
    \resizebox{\linewidth}{!}{
    \begin{tabular}{c c c | c c}
      \toprule
      \multicolumn{5}{c}{\bf CIFAR-10 32 $\times$ 32 } \\
      \midrule
      \bf Rectification & \bf Loss Scaling & \bf Bi-modal Loss & \bf Speed $\uparrow$ & \bf FID $\downarrow$ \\
      \midrule
      \cmark & \cmark & \cmark & 1.63$\times$ & 4.66\\
      \xmark & \cmark & \cmark & 1.49$\times$ & 4.66\\
      \cmark & \xmark & \cmark & 1.62$\times$ & 4.74\\
      \cmark & \cmark & \xmark & 1.63$\times$ & 4.72\\
      \bottomrule
    \end{tabular}
    }
    \caption{Ablation study on timestep-aware loss scaling, knowledge-based model rectification, and the introduction of bi-modal loss.}
    \vspace{-2mm}
    \label{tbl:ablation_3}
\end{table}

\paragraph{Ablation of model loss.} We conduct ablation studies on the loss function, including timestep-aware loss scaling and the introduction of bi-modal loss, to assess their impact on the mask obtained during training. The specific experimental results can be referred to in the third and fourth rows of Tab. \ref{tbl:ablation_3}. We can observe that without the use of timestep-aware loss scaling, the acceleration effect of the mask is diminished, and there is a slight decrease in the quality of the images generated by the accelerated model. When the bi-modal loss is not introduced, the acceleration effect remains consistent with when it is used, but there is a slight decline in the quality of image generation. More experiments can be found in the supplementary material.

\paragraph{Ablation of mask rectification.} We compare the impact of including or excluding knowledge-based mask rectification on the mask, and the specific experimental results can be referred to in the first and second rows of Tab. \ref{tbl:ablation_3}. When knowledge-based model rectification is performed, the quality of the generated images remains, and the acceleration effect increases from 1.49$\times$ to 1.63$\times$. This result validates that our knowledge-based mask rectification can enhance inference speed without sacrificing image quality.

\section{Conclusion}

Our study introduces an innovative approach that significantly accelerates the denoising process of diffusion models with an optimized mask. The essence of this method lies in dynamically adjusting the computational requirements of the model at different timesteps, allowing the model to reduce redundant computations. Moreover, our method does not require additional training data; it only needs Gaussian noise as input for end-to-end training of the mask. Experiments show it works efficiently across DDPM, LDM, DiT, and PixArt models on various datasets, with some even achieving higher image quality post-acceleration. Overall, our work offers new insights into diffusion model acceleration and contributes to the field.

{
    \small
    \bibliographystyle{ieeenat_fullname}
    \bibliography{main}
}

\end{document}